Joseph Jay Williams (williams@cs.toronto.edu), Computer Science, University of Toronto
Jacob Nogas (jacob.nogas@mail.utoronto.ca), Computer Science, University of Toronto
Nina Deliu (nina.deliu@uniroma1.it), Department of Statistical Sciences, Sapienza University of Rome;
MRC - Biostatistics Unit, University of Cambridge
Hammad Shaikh (hammy.shaikh@mail.utoronto.ca), Economics, University of Toronto
Sofia Villar (sofia.villar@mrc-bsu.cam.ac.uk), Biostatistics, University of Cambridge
Audrey Durand (audrey.durand@mcgill.ca), Computer Science, Universitè Laval
Anna Rafferty (arafferty@carleton.edu), Computer Science, Carleton College

## Challenges in Statistical Analysis of Data Collected by a Bandit Algorithm: An Empirical Exploration in Applications to Adaptively Randomized Experiments

### Abstract

Multi-armed bandit algorithms have been argued for decades as useful for *adaptively* randomized experiments. In such experiments, an algorithm varies which arms (e.g. alternative interventions to help students learn) are assigned to participants, with the goal of assigning higher-reward arms to as many participants as possible. We applied the bandit algorithm Thompson Sampling (TS) to run adaptive experiments in three university classes. Instructors saw great value in trying to rapidly use data to give their students in the experiments better arms (e.g. better explanations of a concept). Our deployment, however, illustrated a major barrier for scientists and practitioners to use such adaptive experiments: a lack of quantifiable insight into how much statistical analysis of specific real-world experiments is impacted (Pallmann et al, 2018; FDA, 2019), compared to traditional uniform random assignment. We therefore use our case study of the ubiquitous two-arm binary reward setting to empirically investigate the impact of using Thompson Sampling instead of uniform random assignment. In this setting, using common statistical hypothesis tests, we show that collecting data with TS can as much as double the False Positive Rate (FPR; incorrectly reporting differences when none exist) and the False Negative Rate (FNR; failing to report differences when they exist). We empirically illustrate how and why this occurs–maximizing rewards can lead to biased estimates of arm means (which increases false positives), as well as reduced confidence in estimates of means (which increases false negatives). We find that even when two arms have equal reward, TS can misleadingly allocate many participants to one arm. We show this problem persists using different statistical tests and a Bayesian analysis of the data. We show how two methods for incorporating knowledge of the bandit algorithm into the statistical test can help, but do not eliminate the issues in drawing inferences from adaptive data collection. These empirical results illustrate the nature of the problems to be solved, for more widespread application of bandit algorithms to adaptive real-world experiments, by incorporating statistical considerations into applying, developing, and evaluating bandit algorithms.

# 1. Introduction

Multi-armed bandit algorithms have been put forth for decades as being useful for adaptively randomized experiments (Chernoff, 1966), where an algorithm varies which arms (e.g. alternative interventions to help students learn) are assigned to participants, with the goal of giving higher reward arms to as many participants as possible. For example, recent work has shown bandit algorithms can speed up use of data to help participants in education (Xu et al. 2016; Clement et al., 2015; Williams et al., 2016; Segal et al., 2018), in healthcare (Tewari & Murphy, 2017; Rabbi et al., 2015; Aguilera et al., 2020), and in product design (Li et al., 2010; Chapelle & Li, 2011; Lomas et al., 2016). Yet, these examples are only a tiny fraction of the tens of thousands of experiments where bandit algorithms could be useful, by directing more participants to more effective arms.

In a multi-armed bandit problem, the expected reward for each arm is initially unknown, and the goal is to choose arms in a way that trades-off exploration (learning about the true reward of arms) with exploitation (choosing arms that are more likely to have higher reward, based on the sample's observed rewards from previously chosen arms). Typical bandit algorithms (see Lattimore & Szepesvári, 2020 for an overview) aim to make arm choices that maximize expected cumulative rewards.

Adaptive experiments can be modeled as a bandit problem by considering the allocation of arms (also called conditions, interventions) to participants. At each time point, the algorithm selects an arm for a new participant, and then observes the reward corresponding to the positive or negative outcome for that participant. For example, in an educational experiment, arms might be different versions of hints in an online homework assignment, and the reward might be whether the participant correctly answered a related question after receiving the hint.

A major barrier to adopting bandit algorithms for experimental design is lack of clarity on how statistical analyses of data are impacted when using a bandit algorithm to adapt an experiment (Rafferty, Ying, & Williams, 2019). Theoretical work suggests that adaptive data collection, like that used in bandit algorithms, can induce bias in the estimates of means (Bowden and Trippa, 2017; Deshpande et al., 2018; Nie et al., 2018; Shin, Ramdas, & Rinaldo, 2019; Shin et al. 2020), and that confidence intervals constructed from these statistics may not have correct coverage (Hadad et al., 2019; Zhang et al., 2020). It is therefore necessary to quantify how collecting data using a bandit algorithm impacts the *False Positive Rate (FPR)* of a statistical hypothesis test in a particular application–how often one **incorrectly** concludes the sample data provides evidence for a difference in arm means, when none exists. How is FPR impacted by an adaptive experiment using a bandit algorithm vs a traditional Uniform Random (UR) experiment? Both practical decisions and scientific research rests on knowing and controlling how frequently false positives occur, as these can be deeply problematic (Colquhoun, 2017). Incorrectly concluding one intervention is more effective than another can lead to wasted resources from a practical perspective, and misleads future research that builds on these findings (Ioannidis, 2005; Forstmeier et al. 2017), one of the contributors to a 'replication crisis' in social and behavioral sciences (Open Science Collaboration, 2015; Camerer et al. 2016).

Scientists and practitioners also need empirical insight into the impact on a statistical test's *Statistical Power,* or the probability of concluding there is a difference in arm means when it is truly present. Statistical Power is equal to 1 - False Negative Rate (FNR), where FNR is the probability of failing to conclude a difference exists, when it does. When there are unequal sample sizes and one arm is assigned to fewer participants, lower confidence in the sample estimate of that mean can reduce Power for detecting differences (Villar et al., 2015; Zhang et al., 2020; Yao et al., 2020).

Controlling and quantifying the reduction in False Positive Rate and Power/False Negative Rate is therefore essential, as there is a high bar for scientists and practitioners to trust statistical analysis of adaptively collected data, to ensure best practices are followed and even meet regulatory requirements (FDA, 2019). Combining machine learning and statistics to lower these barriers for analysis of real-world experiments by social-behavioral scientists and practitioners has tremendous opportunity for use of bandit algorithms to have impact.

This paper therefore aims to provide a case study with empirical evaluation into how using bandit algorithms impacts statistical analysis. First, we applied Thompson Sampling a widely used randomized algorithm (Chapelle & Li, 2011; Thompson, 1933), to do real-time adaptation of experiments in three university classes, where instructors saw great value in trying to quickly provide students with better arms, but instructors were concerned about what generalizable conclusions they could draw. We therefore explored how much and why using a bandit algorithm impacts FPR (False Positive Rate) and (Statistical) Power, for a commonly used statistical hypothesis test (Wald test). We target the 2-arm binary reward setting, because it is ubiquitous in experiments, and because if these issues are non-trivial to solve in this case, they will only be more compounded in more complex settings. We constructed and open-sourced a simulation environment with parameterization of arm differences and number of participants inspired by real-world experiments.

We show that when there is no difference in arms (e.g. both arm means have success probability of 0.5), using Thompson Sampling (TS) instead of Uniform Random (UR) can increase FPR (False Positive Rate) from 5% to as much as 13%, which is deeply problematic for applying bandit algorithms to adaptive experiments. Our results explain why and how increased FPR occurs in this setting, and why the problem does not disappear asymptotically. Even though the arm means are identical, reward maximization increases responsiveness to random lows or highs, where a seemingly 'inferior' arm is assigned less frequently and the 'superior' arm more. This causes overly rapidly convergence to one arm, as there is increasing confidence in the (mostly unbiased) mean of the 'superior' arm, but too few samples from the 'inferior' arm to overcome initial 'lows' and identify that its mean is in reality equal to the other arm, resulting in consistently biased estimates of one arm (Shin et al., 2019, Shin et al., 2020).

When there *is* a difference in the arm means (e.g. arm mean of 0.45 vs 0.55), we show Statistical Power of a test to detect this effect is reduced from 80% with UR to 56% with TS. Providing empirical confirmation and illustration of Zhang et al, 2020, Yao et al, 2020, we show that this occurs because putting fewer people into one arm reduces confidence that that arm is truly worse. We show that these problems occur not only for the Wald test, but also for Welch's t-test, and a Bayesian analysis using Bayes factors.

These results can provide guidance as to two ways of getting better statistical analysis of data collected by a bandit algorithm instead of UR. First, to modify or create new statistical tests to use the fact that data was collected using a bandit algorithm (for example, TS varies the probability of assignment). Second, to modify the algorithm or framework for formulating this problem, to be more sensitive to statistical analysis as well as reward. This paper presents illustrative results for how to take the first approach and discusses implications of our empirical results for the second.

We apply and evaluate two methods for adjusting the statistical test. (1) We show Inverse Probability Weighting does successfully reduce bias in the estimates of the means, but only slightly impacts FPR and Power. (2) We evaluate an *Algorithm-Induced* Test, which constructs a non-parametric distribution for hypothesis testing, by simulating data collection using Thompson Sampling. This allows control of the FPR at 5% but greatly reduces Power (to 17%). These illustrate how to combine machine learning and statistics to better analyze data from a bandit algorithm, by modifying the statistical test to incorporate properties of the algorithm.

However, our empirical results suggest the difficulty of this problem, and the need to *also* explore modifications to bandit problem formulations and algorithms to better collect data that allows for reliable statistical analysis. We therefore spell out in the discussion how our work might inform future attempts to develop and evaluate bandit algorithms that are sensitive to these issues, or to formulate alternative formulations/frameworks for the bandit problem. For example, in exploring ways to combine UR exploration and reward maximization, that better balance Reward, FPR (False Positive Rate), and Power.

In summary, our contributions are:

- Highlighting the benefits and barriers to applying bandit algorithms in real-world adaptive experimental designs, through a variety of empirical results:
- Results from deployments of TS (Thompson Sampling) to adaptive experiments in university courses, gaining insight into instructors' perspectives on the benefits and limitations.
- A case-study empirically quantifying how much using TS inflates FPR and and reduces Power, across multiple statistical tests. Our results elucidate how and why a bandit algorithm causes these issues for statistical inference.
- An evaluation of two ways of modifying statistical tests using properties of the bandit algorithm used to collect data. We show how these can improve analysis, but do not fully solve the issues.

Our hope is to illustrate some considerations necessary for applying bandit algorithms to both maximize reward and enable statistical analysis of data, as well as inspire the development and modification of theoretical frameworks and algorithms to better tackle these issues.

## 2. Related Work

### 2.1 Real-world applications of bandit algorithms

Bandit algorithms have been applied to conduct adaptive experiments in different areas where maximizing the immediate user experience takes precedence over generalizable statistical conclusions from that data. This is the classic bandit problem formulation of maximizing *Reward* by selecting arms that have higher reward (or equivalently minimizing *Regret* by not choosing arms that have lower reward). The applications include both industry and research.

Industry settings use bandit algorithms to try to give more popular versions of websites (White, 2012; Hauser et al., 2009), product features or advertisements (Li et al., 2010; Chapelle & Li, 2011; Bakshy et al., 2012), to even find the best available radio channel from a large set of channels (Toldov et al., 2016). However, it should be noted that there are still many product teams that do not use bandit algorithms due to concerns about drawing generalizable conclusions from the data (Kohavi et al., 2012).

Health research has used bandit algorithms to deliver text messages that increase users' physical activity (Aguilera et al., 2020; Murphy & Tewari, 2017; Rabbi et al., 2015), or maximize stress reduction (Paredes et al., 2014). Contextual bandit algorithms have also been applied in health (for a review see Murphy & Tewari, 2017) where the goal is to identify which arm is best, as a function of a particular context vector that captures a user's state or other features. This paper focuses on the non-contextual bandit problem, because if statistical analysis issues are not easily solvable in this setting, they are likely even more heavily compounded in more complex generalizations of the problem, like a contextual bandit.

In education (the application area of our deployment in Section 3), bandit algorithms have been applied to sequencing educational content like courses (Xu et al., 2016) and lessons (Clement et al., 2014; Clement et al., 2015), as well as problem selection (Segal et al., 2018). There have been a few applications of bandit algorithms to adaptive experiments in educational game design (Lomas et al., 2016), evaluating crowdsourced explanations (Williams et al., 2016), and instructional messages (Williams et al., 2018), but they have primarily focused on optimizing learning outcomes rather than questions of how best to analyze the data from the experiments.

These examples show the promise of bandit algorithms, but are the exception rather than the rule in most randomized experiments. To increase the use of bandit algorithms, scientists and practitioners need insight into how to analyze and draw conclusions from the data collected (Pallmann et al. 2018).

## 2.2 Statistical inference from bandit-collected data

There is surprisingly little application of bandit algorithms to adaptive experiments, given the clear relevance, and the fact that clinical trials and online experiments are frequently touted as a key application of bandit algorithms. This is likely due to lack of awareness and certainty as to how statistical inferences can be drawn from adaptive experiments (Pallmann et al. 2018; Burnett et al, 2020). While some theoretical and modeling work in this area exists (e.g., Pallmann et al, 2018), current guidance on adaptive clinical trials emphasizes the need for more applications, case studies, and data sets on which to evaluate the theoretical work. One barrier to the application of these methods is the ongoing debate and concerns about how such adaptive experiments influence properties of statistical hypothesis tests such as False Positive Rates (probability of failing to detect an effect when one exists), and *Statistical Power* (probability of correctly detecting an effect when one exists). For a scientist, a lack of understanding of how much using a bandit algorithm impacts FPR is deeply problematic, as they might draw an incorrect conclusion that a difference exists when none occurs which misleads future scientific research and contributes to the replication crisis (Open Science Collaboration, 2015; Camerer et al. 2016). Even practically, it can waste tremendous practical resources to pursue an intervention that seems promising but is not actually better. In addition, a false negative can be as problematic, when it leads to missing the discovery of an effective intervention.

Drawing inferences from bandit-collected data can be challenging due to biases in the means and other functions of the arms (Atkinson 2014, Bowden and Trippa 2017; Deshpande et al. 2018; Nie et al 2018; Shin, Ramdas and Rinaldo 2019, Shin et al 2020). Thus, some recent work, in the context of hypothesis testing, claims the need for unbiased estimators of means (e.g. Deshpande et al. 2018; Hadad, Hirshberg, Zhan, Wager and Athey 2019; Zhang et al. 2020). This may also solve poor confidence interval coverage (Hadad et al, 2019; Zhang et al. 2020), but the implications for FPR and Power have not been as thoroughly explored. Performance of these estimators relative to one another can vary by sample size (Zhang et al. 2020). In this paper, we build on Bowden and Trippa (2017) work that uses Inverse Probability Weighting (Bowden and Trippa, 2017) to reduce bias; however, this bias reduction inflates the variance of the estimator, which means there are still challenges in drawing conclusions about the relative values of different conditions.

Other work has directly explored the challenges in hypothesis testing with data from adaptive experiments that use bandit algorithms, specifically evaluating FPR and power (Villar et al. 2015; Zhang et al. 2020; Yao et al. 2020). Some of the proposed techniques rely on considerable information about (and control over) the data generating process. For example, recent work considers clipping the probabilities of selecting each action at each point in the data selection (e.g., Yao et al. 2020), or modifying TS so that an action is assigned at most half of the times to encourage more exploration (Kasy and Sautmann, 2020). Some others, by focusing on addressing the biased OLS estimator, require the computation of more advanced estimators, based on adaptive weights and other correlation summaries (Deshpande et al. 2018; Hadad et al. 2019, Zhang et al. 2020). Complementary to this recent theoretical work, rather than developing alternative estimators, as in Zhang et al. 2020, or alternative bandit modifications, as in Yao et al, 2020, our main goal is to provide empirical characterization of the issues in FPR and Power for a specific real-world application.

## 2.3 Balancing reward maximization against other objectives

A theme of the current paper is exploring how reward maximizing bandit algorithms impact another objective – statistical inference from the data collected – and how this is similar to and different from the past work that has considered adaptive assignment of arms to optimize for objectives besides regret. For instance, best-arm identification aims to adaptively assign arms in order to efficiently or accurately identify the optimal arm (Even-dar et al, 2002; Audibert et al, 2010; Russo 2018). Similarly, work on optimal and adaptive experiment designs aims to estimate a parameter of interest with maximum precision and efficiency (e.g., Myung et al. 2013; Kaptein 2014; Smucker et al. 2018).

As algorithms cannot be guaranteed to be optimal on both reward maximization and other objectives, such as best-arm identification (Bubeck et al., 2009), some literature has directly considered the question of how to trade off competing goals. For instance, some work proposed to use a multi-objective bandit problem for trading off cumulative reward against minimizing estimation errors for arm means rewards (Liu et al., 2014; Erraqabi et al., 2017) or to introduce a random cost for pulling an arm and constrain the total cost by a budget (Xia et al., 2015; Hoffman et al., 2014). Other work aims to maximize reward by choosing the best arm, while also gaining enough accuracy in estimating alternative arms to be able to have high confidence that the best arm was chosen and justify generalization about the best arm (Yang et al. 2017; Jamieson & Jain 2018). In this paper, we also consider both reward and an additional objective, related to but different from the previous work. More specifically, our additional goal is to have low false positives and high Statistical Power when testing the null hypothesis of no difference between two arms. We focus specifically on understanding how an interpretable bandit algorithm (Thompson sampling) performs on this alternative objective when collecting data in an adaptive educational experiment. We believe that such empirical evaluations may give useful insights on how an hypothesis testing procedure is affected by the TS algorithm in different plausible settings and how to potentially correct it and ensure stronger guarantees, e.g., in terms of FPR and/or Power. Existing literature has provided only a partial view on this phenomenon, focusing mainly on developing theoretical advances in specific settings.

# 3. Helping Instructors Conduct Adaptive Educational Experiments

To motivate the need for exploring the properties of statistical hypothesis tests on data collected via a bandit algorithm, we present results from adaptive educational experiments conducted in real classrooms (initially reported in Williams et al. 2018).[1] We conducted experiments in three university instructors' courses. Each experiment was conducted in an online system in which students could solve problems to complete required homework. After students attempted a problem, they were randomly assigned a different version of a support message, such as an explanation, hint or learning tip. Each message corresponded to a bandit arm. The support messages were designed in conjunction with instructors to ensure that the comparisons targeted questions that were both educationally relevant and of specific interest to these instructors, making the results more likely to be of use for the instructors' future teaching.

## 3.1 Thompson Sampling with Beta-Bernoulli Bandit

To allocate participants to conditions, we used Thompson Sampling (TS), which is a highly interpretable bandit algorithm that performs well in practice and has received renewed attention in recent years (e.g., Chapelle & Li 2011). TS maintains a posterior distribution over the effectiveness of each arm. To choose one of the $n$ arms $a_1 \ldots, a_n$ at each time step, it samples a value from each of the $n$ posteriors and selects the arm with highest sampled value; this is equivalent to sampling an arm according to the probability that it is the arm with highest mean reward. We denote the probability that a particularly arm $k$ is sampled at time $t$ given the results of previous timesteps as $\pi_{ik}$. After obtaining the stochastic reward $r_t$ at time step $t$, the posterior distribution for the chosen arm is updated. For this application, we use Bernoulli likelihoods to represent the distribution of the reward and the conjugate Beta prior for the success probability, following common practice. The prior for each arm was Beta(19, 1), representing an optimistic prior (i.e., 95% mean prior probability of success) as has been effective in past applications of bandits for selecting educational interventions (Rafferty, Ying, & Williams 2019).

Rewards were based on the student's rating of the helpfulness of the support that they received. In a standard deployment of Beta-Bernoulli TS, each reward would be a single observation of a success or failure, and this standard methodology is what we use in the simulations in section 4 of this paper. However, Chappelle and Li (2011) suggest sharpening or widening the posteriors that are used in sampling to change the relative amount of exploration versus exploitation; in particular, they show that for finite-horizon cases, lower regret can often be obtained on average by dividing the parameters of the posterior distributions by a value $\alpha < 1$, which leads to lower variance in each individual distribution. Because our experiments are conducted in classrooms with a finite number of students and that number of students is often relatively limited, we adopt a version of this technique in the experiments in this section by overweighting both the prior and each student rating by a factor of 10 in the posterior, similar to setting $\alpha = 0.1$. Each student gave a rating $x$ between 0 and 10 for the helpfulness of the support they received, and to update the posterior, this was treated as 10*($x$/10) successes and 10*(1- $x$/10) failures in the posterior update. We consider the consequences of this choice in the analysis of the results by calculating the posterior probability an arm is optimal given a more standard approach with a Beta(1,1) prior and no overweighting of ratings.

---

[1] This is a summary of previously collected data from Williams et al, 2018, with some additional analyses and exposition of the data collection issues that are particularly relevant for a machine learning rather than human-computer interaction audience, focusing specifically on the issues in statistical analysis which the original paper did not address.

One of the reasons we chose TS was its interpretability, as instructors wanted to know what was happening in the experiments in their classes. Instructors could view a data and policy dashboard at any time to examine the results of the experiment and see the probability that TS would assign the next student participant to each arm. The assignment probability provides an easy way for instructors to see both what arm (i.e., type of support) the system estimates to be most helpful and the degree to which the system is deviating from even allocation.

Following data collection, we compare the effectiveness of the arms using the Wald test, a common statistical hypothesis test. For Experiments 1 and 3, which include two arms, we test the hypothesis that the mean effectiveness of the two arms differs, and for Experiment 2, which includes three arms, we test that each of the more elaborate supports (hint and worked solution) differ from a minimal support (answer only). This test relies on a Maximum Likelihood Estimator (MLE) of the effectiveness of each arm, which we denote $\hat{p}_k^{MLE}$ for arm $k$, and calculate as

$$\hat{p}_k^{MLE} = \frac{\frac{1}{n}\sum_{i=1}^{n} r_i \delta_{ik}}{\frac{1}{n}\sum_{i=1}^{n} \delta_{ik}}$$

where $\delta_{ik}$ is 1 if student $i$ was assigned to arm $k$ and 0 otherwise. This estimate ignores the fact that the data was collected by a bandit, which we consider and address in later sections.

**Table 1:** *Results of three adaptive educational experiments where data was collected using Thompson Sampling. Instructors saw the assignment probability, arm mean, and total number of assigned students while the experiment was running, providing real-time information about the experiment. Column 4 includes the posterior probability each arm is optimal according to TS when a beta(1,1) prior is used and there is no overweighting of ratings. The posterior probability an arm is optimal according to TS differs from assignment probability due to the overweighting of each individual student in this experiment. [a]Represents the probability of the next student being assigned to each arm at the end of the experiment. [b]For experiment 2 the reference group is "Answer" and is compared against the "Hint" and "Solution" groups using the statistical hypothesis test. [c]The last column presents the Wald test statistic and p-value for the hypothesis testing of whether this is a significant difference in the arm means.*

| Experiment | Arm | Assignment Probability at end of experiment[a] | Posterior. Prob. Arm is Optimal | Mean (SE) | Total assigned students | Wald Test Statistic[c] [p-value] |
|---|---|---|---|---|---|---|
| 1) Type of explanation | Quantitative | 0.23 | 0.4 | 0.73 (0.066) | 46 | 0.2509 [0.802] |
| | Analogical | 0.77 | 0.6 | 0.75 (0.058) | 56 | |
| 2) Type of help[b] | Answer Only | 0.00 | 0.04 | 0.25 (0.306) | 2 | NA |
| | Hint | 0.00 | 0.10 | 0.58 (0.201) | 6 | 0.9088 [0.364] |
| | Worked Solution | 1.00 | 0.86 | 0.79 (0.053) | 60 | 1.7316 [0.083] |

| 3) Type of feedback prompt | Feedback | 0.48 | 0.46 | 0.90 (0.071) | 18 | 0.1023 [0.919] |
|---|---|---|---|---|---|---|
| | Feedback and Prompt to Review | 0.52 | 0.54 | 0.91 (0.067) | 18 | |

## 3.2 Results: Instructor perspectives

We conducted interviews with all three instructors to collect information about their experiences using the system in their classes. Instructors found it valuable to have data be more rapidly used to help future students, and were more willing to conduct adaptive experiments in the future due to the potential for current students to benefit, rather than the experiment being solely for academic knowledge. Instructors appreciated that they could use the dashboard to observe the assignment probability throughout the adaptive experiment, pointing to the usefulness of an interpretable algorithm like TS. They also wanted to be able to encode prior knowledge in future versions of the system, an important focus area for future work.

## 3.3 Results: Data collection and inference

As shown in Table 1, the three experiments had quite different patterns of adaptation: one in which the mean effectiveness seemed to be similar across arms and the probabilities of assignment were also similar (Experiment 3), one in which the mean effectiveness was somewhat similar across conditions but the probabilities of assignment were more substantively different (Experiment 1), and one in which the arms were relatively different and the probabilities of assignment were extremely different, with almost all probability on one condition (Experiment 2). Here, probability of assignment for each arm refers to the probability that TS will assign the next student to each arm. From these results, one might be tempted to conclude that there is no real difference between the conditions in Experiment 3, while there is a difference between the conditions in the other two experiments.

However, the results of the statistical hypothesis tests in Table 1 do not suggest this inference: in all cases, the test statistic suggests there is not sufficient evidence to conclude that the arms differ in effectiveness. The sample sizes here are relatively small, which means that even an experiment using UR allocation might have a low probability of detecting differences among conditions. Yet, the highly uneven allocation suggests that even with more students, we would be unlikely in cases like Experiment 3 to collect sufficient evidence to conclude with high confidence that the arms differed (if in fact they do). One might attribute this skew to the overweighting of each individual response, which was intended to make TS more exploitative. However, even when considering the posterior probabilities from traditional TS (column 4 in Table 1), it seems likely that an instructor would believe that the arms differ from one another in Experiment 2, despite this conclusion not being supported by statistical hypothesis testing. This suggests that even without heightened exploitation, adaptation can be problematic for drawing conclusions: the adaption in Experiment 2 led to over 88% of students being placed in a single arm, and the small numbers of students in the other arms led to high uncertainty about the true effectiveness of these alternative arms, as represented by large standard errors of the mean. T; this is expected behavior from TS, but is in tension with the goal of drawing conclusions about arm differences (see, e.g., Liu, Mandel, Brunskill, & Popovic 2014).

Determining whether or not arms (i.e., educational interventions) differ is of central importance for instructors deciding whether to make pervasive changes in their courses and for researchers who are building a body of knowledge. As previously discussed, prior work has suggested that drawing inferences from bandit-collected data can be challenging (e.g. Bowden and Trippa, 2017; Rafferty et al. 2019; Villar et al. 2015), and these results from a real deployment of TS for educational research provide an example of the kind of deployment that instructors are receptive to, but where solutions are needed for how to draw generalizable conclusions. In the following section, we investigate the issues of statistical hypothesis testing from bandit-collected data to determine both what issues occur and how effectively they can be addressed with small changes to the testing procedure. If small changes to the testing procedure are insufficient, it suggests the need for machine learning work that modifies the algorithms themselves or formulates alternative frameworks that balance reward with statistical analysis.

## 4. Challenges in Analyzing Data Collected by Bandit Algorithms

This section examines and quantifies how using a bandit algorithm (like Thompson Sampling) to collect data impacts the statistical analysis, such as the level of confidence in the final experiment conclusion of whether or not there is truly a difference in arms. We therefore compare traditional experiments with UR (Uniform Random) collected data against adaptive experiments where data is collected by a bandit algorithm. We focus on the widely used TS (Thompson Sampling), but also include a comparison to an Epsilon-Greedy bandit algorithm (EG or Epsilon-Greedy) to demonstrate how different types of adaptivity differentially impact Power and FPR. We consider the case study of ubiquitous 2-arm experiments with binary rewards, as statistical analysis issues arising in this fundamental setting may be only compounded with more complex experimental designs. To evaluate statistical analysis of data from the UR, TS, and EG data collection strategies across many repeated applications to conduct a randomized experiment, we construct a simulation environment informed by real-world experiments and past literature. We focus first on Wald's test, a statistical hypothesis commonly used in practice, and demonstrate decreased Power and increased False Positive Rate (FPR) for TS relative to UR, while EG has FPR close to UR but at the cost of severely reduced Power. We then examine the impact of both alternative hypothesis testing strategies, exploring Welch's test to correct for unequal sample sizes in the arms and a Bayes factor analysis, demonstrating that the same issues occur even with these modifications.

We empirically explore why using a bandit algorithm to collect data problematically increases the FPR. To preview what we find: When there is no difference in arm means and no reward to be gained, TS responds to sampling variability that makes one arm look 'lower' than the other and stops assigning it, leading to a bias in estimating the 'lower' arm but high confidence in the 'higher' arm's mean. We also illustrate how Power is reduced (FNR increased): When one arm mean is higher, assigning the lower arm to few participants means there is little confidence in its estimate, and therefore reduced certainty that the difference in sample means is due to a true difference, versus noise. Since TS tends to converge on giving one arm to most participants (and this does not change even with large samples), having more participants in one arm provides less evidence for that arm being higher/superior than might be expected. These results illustrate the challenges to be surmounted in adaptive experiments, such as in modifying statistical tests to incorporate knowledge of the dynamics of data collection, or modifying how the bandit algorithms collect data.

### 4.1 Methods: Simulation Studies Based On Our Motivating Application

**Simulation Environment, True Arm Difference, and Experimental Design.** For our two arm setting with a binary reward outcome, we constructed a simulation environment that allows for varying the values and differences in arm means, as well as the sample size or number of trials. We provide this along with our implementation of the algorithms and statistical tests as open-source code, so that others can replicate our results and evaluate other bandit algorithms and statistical tests[2].

   **True Arm Differences and Sample Size:** We examine cases where the arms have equal rates of rewards ( $p_1 = p_2 = 0.5$ ) and where there is a difference of 0.1 in the reward rate ( $p_1 = 0.55$, $p_2 = 0.45$ ). The latter corresponds to a small effect size, as measured by Cohen's $w$ (Cohen 1988); in many cases, effect sizes are small in experiments in education and other social/behavioral sciences. In all simulations, we use a sample size *n*=785 simulated participants. This is the sample size needed for UR-collected data to have 80% Power given the true arm differences that we use.

   **Data Collection Strategies (UR, TS, EG).** To simulate conducting an experiment, the reward for each participant is generated from the reward distribution of the chosen arm, where the assignment of arms was conducted using (1) TS, (2) UR, or (3) EG. TS used a conjugate Beta-Bernoulli model with each arm having a prior of Beta(1,1). EG used $\varepsilon$ = *0.1*, so that for each participant/trial, with probability $\varepsilon$ = *0.1* arms were assigned using UR, and with probability 1- $\varepsilon$ = 0.9 greedy assignment of arms assigned the arm with the highest sample mean at that point in time. TS and EG each had a batch-size of 1, so that the assignment probability (for TS, posterior probability of an arm having the highest arm mean) was updated immediately after observing the reward for every participant, before assigning the next participant.

   **Simulation design.** We conducted 5000 simulations for each assignment strategy (TS, UR, and EG), and each arm difference {0, 0.1}, resulting in 5000 x 3 x 2 = 30 000 simulations. This simulated the application of UR/TS/EG to 5000 randomized experiments with a particular experimental design (sample size = 785, number of arms = 2, binary reward) and two different environments (or data generating processes) of arm difference of 0 vs 0.1.

   **Hypothesis Testing.** As is standard practice for scientists/practitioners in hypothesis testing, we conducted our hypothesis test to obtain a fixed FPR from UR-collected data. Specifically, we aimed for an FPR of 5% in a two-sided test (we tested whether arm 1 was greater than arm 2, or arm 2 greater than arm 1). This is done by choosing the *Critical Value* of the Test Statistic so that, when there is no difference in arm means, the probability of a particular sample producing a test statistic as extreme as the critical value is 5% or less. For example, as shown in Figure 3, the Wald test statistic *Critical Value* was 1.96, meaning that given no difference in arm means, the probability of a test statistic >1.96 or < -1.96 was 5% for UR-collected data. For each assignment strategy (UR, TS, and EG), we computed Power and FPR. FPR is computed as the proportion of the 5000 simulated experiments with arm difference 0 where the statistical hypothesis test concluded there was a difference in arm means (e.g. value of the Wald test statistic from the sample was ±1.96 or greater, p-value was < 0.05). Power was computed as the proportion of the 5000 simulations where arm difference was 0.1 in which the statistical hypothesis test concluded there was a difference in arm means (recall that FNR (False Negative Rate) = 1 - Power). We primarily focus on the Wald test statistic, which is defined as:

---

[2] https://github.com/IAIresearchlab/SCBandits/tree/master/BanditsInferenceJosephJacob

$$\frac{\hat{p}_1 - \hat{p}_2}{\sqrt{\frac{\hat{p}_1(1-\hat{p}_1)}{n_1} + \frac{\hat{p}_2(1-\hat{p}_2)}{n_2}}}$$

where $\hat{p}_k$ is an estimate for the mean reward for arm k, such as from the MLE estimator shown in Section 3.1. We use the Wald test as it is commonly employed by practitioners who conduct balanced AB tests, and previous work has suggested it is also a reasonable choice for response-adaptive trials (Hu & Rosenberger, 2006). Under some adaptive designs, it was demonstrated to be asymptotically optimal in terms of achieving the upper bound of the asymptotic power (Yi and Li, 2018), with improved performances in statistical power for small to moderate sample sizes, compared to both score and likelihood ratio tests (Yi and Wang, 2011). We also compared performance of this test statistic to two alternative approaches: the frequentist Welch's t-test and an approach based on the Bayes factor. Methods for these approaches are found in the Appendices.

## 4.2 Drawing Statistical Conclusions from TS-collected vs UR-Collected data

Table 2 shows the FPR and Power, using different statistical tests, for data collected using TS, UR, and EG. As shown in Row 1 of Table 2, the FPR is higher for TS than UR using the Wald test. When there is no difference in arm means, the FPR is controlled at 5% for UR, but this goes up to 13% for TS.

Statistical Power is also reduced. With the Wald test, we achieve a Power of only 56% for TS (FNR of 44%) compared to 80% for UR (20% FNR). This might make an instructor reluctant to use a bandit algorithm, as the chance that a statistical analysis correctly detects a difference when it is present decreases from 80% with UR to 56% with TS.

We also see that using Jeffreys' prior does not influence results substantively, suggesting consistency of results across different priors; see Appendix 2 for more details.

**Table 2:** *FPR (False Positive Rate, probability of reporting a difference when arm means equal) and Power (probability of correctly reporting a difference when arm mean difference is 0.1) for analysis of data collected using Thompson Sampling (TS), Uniform Random (UR), and Epsilon Greedy (0.1). Each row shows FPR and Power for a different Statistical Analysis, including: (1) Wald test, (2) Welch's t-test used for unequal sample sizes or variances, (3) Bayes factor (with cutoffs of 0.4, 1.0, and 3.0). (4) IPW-adjusted Wald test: Wald test where the sample means are adjusted using the TS Assignment Probabilities to do Inverse Probability Weighting (IPW). (5) TS-induced Wald test: Wald test where a non-parametric distribution is used that is generated by simulations of running TS under the null hypothesis that arm means are equal. The Standard Errors (SE) are written in parentheses.*

| Statistical Test | Thompson Sampling | | Uniform Random | | Epsilon-Greedy (epsilon = 0.1) | |
|---|---|---|---|---|---|---|
| | FPR (SE) [ $n = 785$ ; $p_1 = p_2 = 0.5$ ] | Power (SE) [ $n = 785$ ; $p_1 = 0.55$ $p_2 = 0.45$ ] | FPR (SE) [ $n = 785$ ; $p_1 = p_2 = 0.5$ ] | Power (SE) [ $n = 785$ ; $p_1 = 0.55$ $p_2 = 0.45$ ] | FPR (SE) [ $n = 785$ ; $p_1 = p_2 = 0.5$ ] | Power (SE) [ $n = 785$ ; $p_1 = 0.55$ $p_2 = 0.45$ ] |
| 1. Wald test | 13.4 % (0.5) | 56.2 % (0.7) | 5.0 % (0.3) | 81. 2% (0.6) | 6.3 % (0.3) | 39.2 % (0.7) |
| 2. Welch's t-test | 11.9 % (0.5) | 52. 4% (0.7) | 5.0 % (0.3) | 81.2 % (0.6) | 5.4 % (0.3) | 36. 8% (0.7) |
| 3. Bayes factor (Cutoff 3.0) | 4.2 % (0.3) | 19.3 % (0.6) | 0.5 % (0.1) | 49.8 % (0.7) | 1.2 % (0.2) | 17.5 % (0.5) |
| 4. Bayes factor (Cutoff 1.0) | 9.3 % (0.4) | 43.5 % (0.7) | 1.7 % (0.2) | 66.5 % (0.7) | 3.9 % (0.3) | 32.4 % (0.7) |
| 5. Bayes factor (Cutoff 0.4) | 18.6 % (0.6) | 69.5 % (0.7) | 5.0 % (0.3) | 81.3 % (0.6) | 12.0 % (0.5) | 51.7% (0.7) |
| 6. IPW-adjusted Wald test | 11.0 % (0.4) | 35.9 % (0.7) | | | | |
| 7. TS-induced Wald test | 4.5 % (0.3) | 17.4 % (0.5) | | | | |

### 4.2.1 When there is No Difference in arm means, False Positive Rate is increased by biased underestimation of a 'lower' arm and increasing confidence in the 'higher' arm.

As discussed in the previous section, when the arm difference is 0 (e.g. both arm means have success probability of 0.5), using TS instead of UR can increase FPR from 5% to as much as 13% with the Wald test, which is problematic for false discoveries. Figure 1 shows one of the simulations of TS for Arm Difference = 0.0 that illustrates why TS inflates FPR compared to UR. When there is no true underlying difference in arm means, sampling variability nevertheless will lead to sample estimates of means that make one arm look 'lower' than the other (the sample mean is lower, but the true mean is not). Since TS is adaptive to the posterior probability (which is sensitive to the sample mean), when (by chance) an arm's sample mean is lower than its true mean, TS will assign fewer participants to it and assign them to the other 'higher' arm which has a sample mean closer to (or higher than) its true mean. Increasingly assigning the seemingly 'superior' arm to participants reduces the standard error in the estimate of its sample mean and increases confidence in the value of that 'higher' arm, even though in reality it is not higher, the 'lower' arm simply has a sample mean that is below its true mean.

Even as more and more data is collected, TS does not substantially correct the earlier biased estimate of the 'lower' mean. Figure 1 illustrates how the assignment probability only gets higher and higher as confidence in the higher mean increases, and this convergence means there is increasingly less assignment of the "lower" arm, and so little opportunity to obtain data that contradicts earlier misleading observations.

Notably, these results suggest that the convergence of a bandit algorithm to one arm is not in itself strong evidence that there is actually a difference in arm means. The adaptation of a reward-maximizing algorithm to random lows and highs can result in a very high assignment probability (and far more participants) being assigned to one arm. TS is more likely than UR to result in one arm having a higher sample mean than the other, even when the true means are identical.

**Figure 1:** *This figure shows an illustrative example of TS behaviour over the course of a single simulation experiment, which results in a false positive. This is a scenario where the true arm difference is 0 with p1 = p2 = 0.5 (mean reward for arm 1 and arm 2) and sample size n = 785. The sample mean estimates of arm 1 are displayed in blue, and the sample mean estimates of arm 2 are displayed in orange (left vertical axis). The assignment probability for arm 1 is denoted in green (right vertical axis). The vertical bars around the sample means represent 95% confidence intervals.*

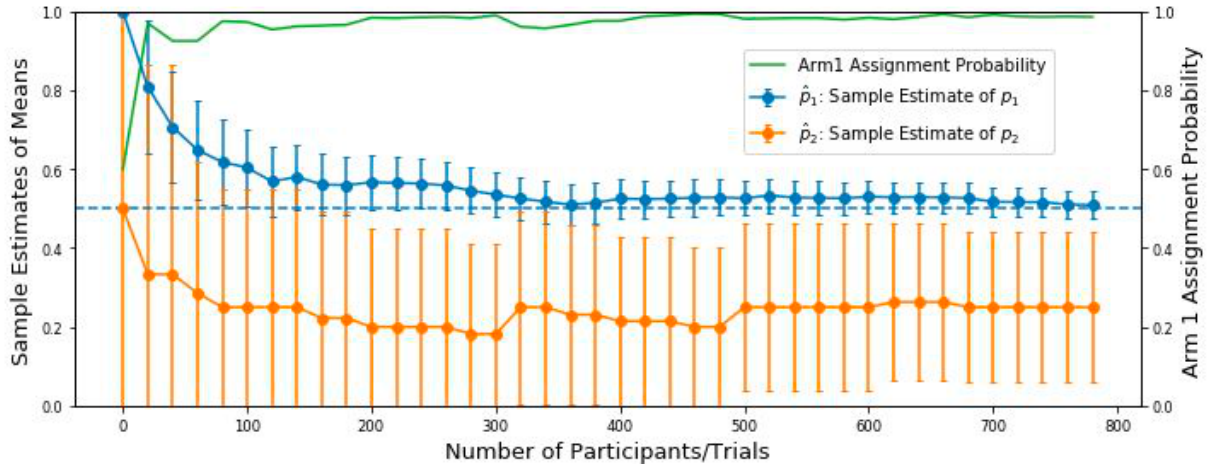

**Figure 2:** *This figure shows an illustrative example of TS behaviour over the course of a single simulation, which results in a false negative. This is a scenario where the true arm difference is 0.1 with p1 = 0.55 (mean reward for arm 1), p2 = 0.45 (mean reward for arm 2) and sample size n = 785. The sample mean estimates of arm 1 are displayed in blue, and sample mean estimates of arm 2 are displayed in orange (left vertical axis). The assignment probability for arm 1 is denoted in green (right vertical axis). The vertical bars around the sample means represent 95% confidence intervals.*

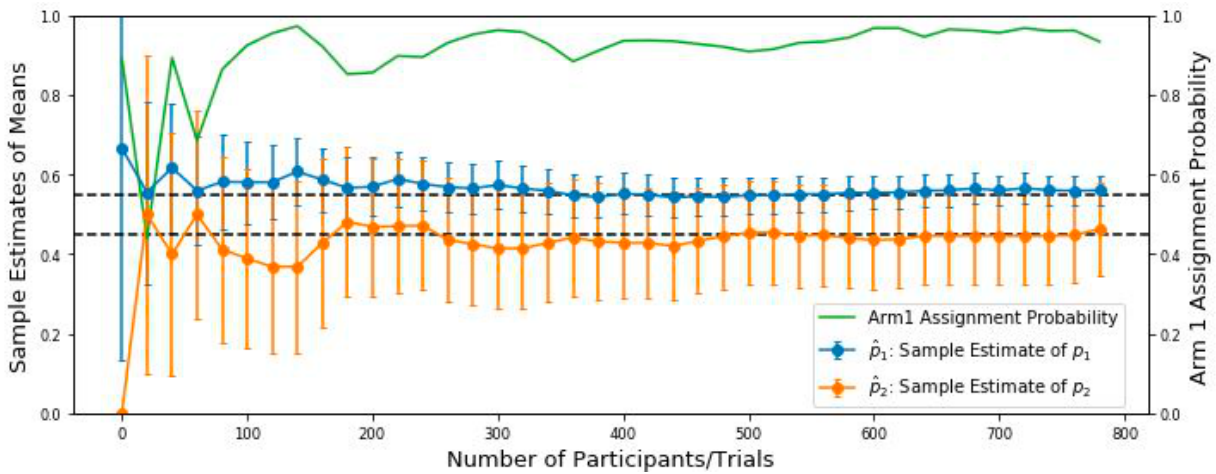

**Figure 3:** *Distribution of the Wald test statistic for Arm Difference of 0.0 and 0.1, for (a) Uniform Random (UR) sampling, and (b) Thompson Sampling (TS). This uses a Critical Value for the Wald Test Statistic of 1.96 for hypothesis testing. The FPR is shown as the red areas under the curve, and Power as the green area. This Figure shows that the FPR is increased when using TS (vs UR) because the distribution of the Wald statistic has a larger variation under arm difference = 0. Using TS (vs UR) decreases the power because the mean of the Wald statistic distribution is smaller under arm difference = 0.1, because of the reduced sample size and therefore reduced confidence in the mean of arm 2.*

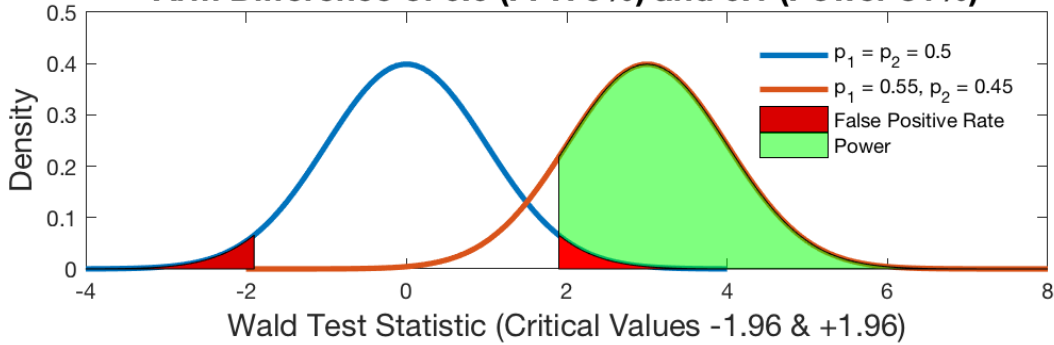

**(a) Wald Test Statistic Distribution under Uniform Random Assignment: Arm Difference of 0.0 (FPR 5%) and 0.1 (Power 81%)**

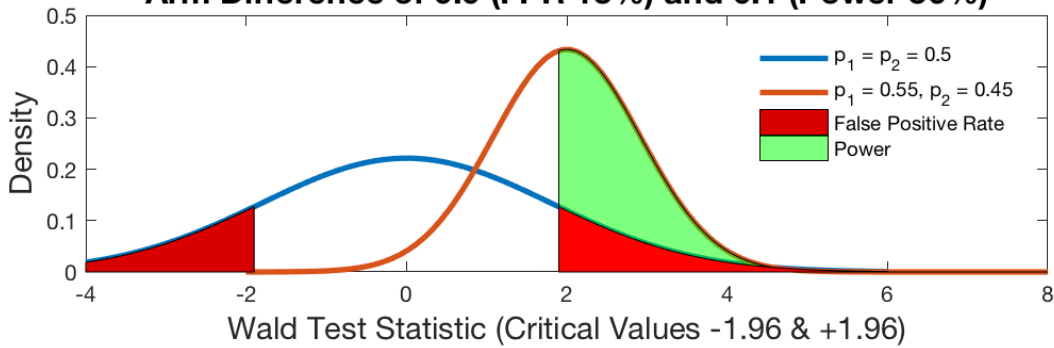

**(b) Wald Test Statistic Distribution under Thompson Sampling Assignment: Arm Difference of 0.0 (FPR 13%) and 0.1 (Power 56%)**

*4.2.2 Power to detect a difference is reduced because unequal assignment leads to reduced confidence in estimate of lower sample mean*

When there is a difference in the arm means (e.g. arm mean of 0.55 vs 0.45), we showed Statistical Power to detect this effect was reduced from 80% with UR to 56% with TS (see Table 2). This means that trying to improve the reward for participants in an experiment takes the FNR from 20% to 44%, more than doubling how often an experiment might fail to report a difference, when it exists. The Power decreases since TS assigns very few samples into the truly worse arm, decreasing the overall confidence that a true difference exists. Although TS underestimates the worse arm, which increases the estimated difference, this is outweighed by the far larger increase in the standard error (and confidence interval) for the sample mean. This is shown in Figure 2, as TS allocates most of the participants into the better arm, increasing the reliability of the estimate as shown by the confidence interval becoming tighter around the mean. However, great uncertainty persists in the sample mean for arm 2, illustrated by the confidence intervals remaining fairly large even after the experiment is complete. Since the inflation in the estimated difference is outweighed by the corresponding change in the standard error, this results in a smaller Wald test statistic under the alternative, and decreases the Power, as illustrated in Figure 3.

### 4.2.3 Higher Assignment probability is not sufficiently indicative of a difference.

One might hypothesize that when TS has a higher assignment probability for one arm and puts more participants in one arm, this higher posterior probability that the arm is optimal could in itself be more reliable evidence for a difference in arms, than the statistical hypothesis tests we have investigated. However, we see evidence against this idea, and reason to see high assignment probabilities as providing less evidence for a difference than might be presumed. The results in Table 3 show that even when there is no difference in arm means, the assignment probabilities can mislead one to believe that an arm is superior. For example, when the difference in arm means is 0, the assignment probability will be 0.7 or higher **69%** of the time. When the difference in arm means is 0.1, the assignment probability will be 0.7 or higher 98% of the time.

Table 3 shows what proportion of the 5000 simulations result in a particular Assignment Probability, organized by the Assignment Probability of the "Superior" Arm – whichever arm has the higher assignment probability in a simulation. So a simulation where arm 1 had an assignment probability of 0.74 would contribute to the Proportion of Simulations in the interval [0.7, 0.8], as would a simulation where arm 2 had an assignment probability of 0.76.

These results show that large disparities in the assignment probabilities across arms (e.g. arm 1 is 0.2 and arm 2 is 0.8) are weaker evidence for a difference actually existing, than might be hoped. Even when there is no difference in arm means, the assignment probabilities can get very high, quite often – assignment probabilities of 0.9 or higher occur 33% of the time when there is **no** difference in arm means, as TS tends to converge on one arm. Therefore when the arm difference is 0, the posterior probabilities are unreliable for inferring whether a real difference across the arms exists due to adaptive policies being sensitive to random noise as a result of their reward maximizing nature.

**Table 3.** *Distribution of Assignment Probability when Arm Difference is 0 (column 2) and 0.1 (column 3). This Table shows the proportion of simulations in which an arm has a certain assignment probability at the completion of an experiment. To illustrate that assignment probabilities are still extreme when there is no difference in arm means, we bin by the "Superior" Arm (whichever arm has Higher Assignment Probability in a simulation, regardless of what the true arm means are) at various assignment probability intervals (column 1)*

| Assignment Probability of "Superior" Arm (Higher Assignment Probability) | Percentage of Simulations [Arm Difference 0.0] | Percentage of Simulations [Arm Difference 0.1] |
|---|---|---|
| [0.5, 0.6] | 15% | 1% |
| [0.6, 0.7] | 15% | 1% |
| [0.7 ,0. 8] | 16% | 3% |
| [0.8, 0. 9] | 20% | 8% |
| [0.9, 1.0] | 33% | 87% |

### 4.3 Alternative bandit strategy: impact of Epsilon-Greedy on FPR and Power

The results presented so far suggest the relevant FPR and Power issues would arise for other bandit algorithms, because of how reward-maximization changes responsiveness to random highs and lows. Of course, the exact pattern of how much Power is decreased and FPR is increased could vary, and this is a key direction for future research.

Here, we investigated the Epsilon-Greedy algorithm, with epsilon of 0.1, or 10% of participants assigned using UR, the remainder greedily to the arm with higher sample mean. One might predict EG would perform well due to explicitly incorporating UR assignment. When using the Wald test, Table 1 shows that EG seems to have better FPR compared to TS (6% vs 13%), which is also quite close to the 5% of UR. However, this is misleading, as Power is reduced to only 39%, compared to 56% for TS and 81% for UR. Because of greedy assignment, far fewer participants are assigned to the 'lower' arm and the very small sample sizes mean there is insufficient evidence for a statistical test to conclude there is a difference.

In addition to EG's highly reduced Power, EG fails to attenuate its adaptiveness in cases where there is less certainty about the difference, where for the sake of both reward and inference, it would be preferable to do more exploration.

### 4.4 Alternative hypothesis testing approaches: Welch's t-test and Bayes factor do not solve the FPR and Power issue

#### 4.4.1 Analysis using Welch's t-test shows similar results as the Wald test

While the prior results suggest that properties of the data itself, not specifics of the Wald test, lead to the issues with FPR and Power, one might hypothesize that the issues could be resolved using a test better suited to experiments with unequal sample sizes. We therefore consider Welch's t-test, which is typically used for handling unequal variances and/or unequal sample sizes in arms, and so might be particularly suitable for an adaptive experimentation based on bandit algorithms. However, Welch's test does not correct the issues: as shown in Table 2, the FPR is only 1.5% less than using Wald test, and this reduction comes with a reduction in Power of 3.8%. Details of the Welch's t-test are given in Appendix 4.

This illustrates that the issue is with biases in estimates of means and reduced confidence in those estimates, as illustrated earlier. The Welch's t-test and Wald test are asymptotically equivalent (i.e., as the sample size, $n\rightarrow\infty$, they will reject the same cases). The advantage of Welch's t-test for handling unequal sample sizes (and therefore unequal variance in the estimate of the sample mean) is largely observed in small sample sizes, so we would not expect a huge difference in performances with a sample size similar to the one we have in our experiments (n = 785, smallest is n = 88) as typically a heuristic is that the advantage of t-tests over z-tests rapidly diminish from a sample size of n=30 and larger (Hogg et al., 2010; Casella & Berger, 2002).

#### 4.4.2 Analysis using Bayes factor shows same patterns as Wald test

While both frequentist tests exhibit the same issues with FPR and Power, one might hypothesize that these issues would be eliminated or mitigated by using a Bayesian framework. A Bayesian analysis compares two hypotheses as a special case of model comparison, providing a measure for which model (i.e., hypothesis) better fits the data and quantifies the strength of that support. We perform a Bayes factor analysis comparing the evidence in favor of the alternative hypothesis that there is a difference in arm means with the null hypothesis that there is not a difference between arm means. Details on how we computed the Bayes factor in the binary reward setting and prior choices are in Appendix 1. We consider two cutoffs as the "critical" values for selecting the alternative hypothesis over the null hypothesis: > 1, which means that the evidence favors the alternative hypothesis over the null hypothesis, and > 3. The choice of 3 choice is based on Jeffreys' scales of evidence for model selection (Jeffreys, 1961; Kass & Raftery, 1995), which considers a Bayes factor > 3 as substantial evidence in favour of one hypothesis or the other. While this approach is a combination of Bayesian and frequentist analysis methods, rather than purely Bayesian, its results can help to illustrate that the issues with frequentist hypothesis testing in this setting are not caused only by idiosyncrasies of the tests we examined.

When using this Bayes factor approach, as shown in Table 2, we see that the test is overall significantly more conservative than the Wald test even for UR sampling: using the cutoff of 3 gives a FPR of only 0.5% and Power of 49.8%. Relative to these values, TS sampling increases the FPR to 4.2%, while further decreasing the Power to 19.3%. The looser cutoff of 1 shows the same where 1 indicates equal support for both models is somewhat less conservative, but the same trends of inflated FPR and decreased Power occurs when comparing the results from data collected via TS rather than UR. This demonstrates that pattern of results when evaluating the data using a Bayesian-inspired approach mirrors the pattern of the purely frequentist methods.

## 5. Adjusting existing statistical tests to incorporate knowledge about the adaptive data collection process induced by the bandit algorithm

We now examine two approaches based on the literature to address the inflated FPR and reduced Power problem discussed in the previous sections. These approaches are similar in the sense that they both adjust the Wald test statistic by incorporating knowledge about the data generating process that is induced by the adaptive nature of the bandit algorithm. However, while the first proposal relies on replacing the biased MLE used in the Wald test formula, the second proposal directly estimates the distribution of the test statistic in order to derive adjusted Critical Values.

Most of the statistical tests, including the parametric Wald test, are based on some underlying assumptions of the data, such as the iid assumption. In a traditional UR experiment, where data are iid, the MLE has strong theoretical properties such as unbiasedness and normality distribution. However, in adaptive experiments, an extensive number of studies have demonstrated both that the unbiasedness property does not hold anymore (Bowden and Trippa 2017; Deshpande et al. 2018; Nie et al 2018; Shin, Ramdas and Rinaldo 2019, Shin et al 2020) and that the Wald test does not satisfy the theoretical standard normal distribution (Jamieson & Jain 2018; Hadad et al, 2019; Zhang et al, 2020). We investigate two alternative hypothesis testing procedures that address these issues: an *IPW-adjusted Wald test* and a *TS-induced Wald test* (based on a previous work of Smith and Villar, 2018).

With the IPW-adjusted Wald test we propose replacing the biased MLE in the Wald test formula with an unbiased estimator in order to evaluate whether correcting the bias might also improve the FPR and Power. The unbiased estimator we use here is the Inverse Probability of Weighting (IPW) estimator, proposed first, in causal inference literature (Robins et al, 1994; Robins, 2000), and then, adopted by Bowden and Trippa (2017) in the context of adaptive clinical trials with data collected by Play-the-winner strategy. In the context of an educational experiment if we now denote with $n$ the total number of students and with $r_i$ the binary outcome for the $i$-th student, the IPW estimator is given by:

$$\hat{p}_k^{IPW} = \frac{\frac{1}{n}\sum_{i=1}^{n} r_i \frac{\delta_{ik}}{\pi_{ik}}}{\frac{1}{n}\sum_{i=1}^{n} \frac{\delta_{ik}}{\pi_{ik}}},$$

where $\pi_{ik}$ is the randomization probability for student $i$ to version $k$, with $k = \{1, 2\}$, and $\delta_{ik}$ is the delta function, which takes value *1* if student $i$ is assigned to version $k$ and *0* otherwise.

Intuitively, the IPW estimator for arm $k$ is a weighted average of observed rewards from arm $k$, i.e., $\delta_{ik} r_i$, with weights given by the inverse of the probability for that participant to be assigned to condition $k$, i.e., $1/\pi_{ik}$: the higher the probability of receiving a specific treatment, the lower the weight of the reward. This allows a fairer comparison across arms with different sample sizes. By replacing now the biased MLE in the Wald test formula with the unbiased IPW estimator we obtain the proposed IPW-adjusted Wald test given below:

$$\frac{\hat{p}_1^{IPW} - \hat{p}_2^{IPW}}{\sqrt{\frac{\hat{p}_1^{IPW}(1-\hat{p}_1^{IPW})}{n_1} + \frac{\hat{p}_2^{IPW}(1-\hat{p}_2^{IPW})}{n_2}}},$$

with $n_1$ and $n_2$ denoting the sample size of arm 1 and 2, respectively.

The second alternative we investigate is a TS-induced Wald test. Rather than addressing the biased estimator, this approach addresses the asymptotic distribution of the Wald test distribution. While the Wald test follows a standard normal distribution in the null case given iid data, but as we showed in Section 4, this distribution is not obtained with TS-collected data. We thus propose a more flexible non-parametric approach (Smith and Villar 2018), by estimating the empirical distribution of the Wald test that is induced by the TS assignment procedure under the null, i.e., when there is no difference in arm means. We then derive its adjusted Critical Values as the empirical quantiles of the estimated TS-induced Wald test distribution at the desired significance levels of $\alpha/2 = 0.025$ and $1 - \alpha/2 = 0.975$ for obtaining a 5% FPR control.

## 5.1 Performance of the IPW-adjusted Wald test

The IPW-adjusted Wald test approach aims to explore the impact on FPR and Power when the IPW unbiased estimator rather than the biased MLE is plugged in the Wald test formula.

**IPW reduces bias in sample estimates of arm means.** First, by comparing the IPW estimator with the MLE estimator we can gain insights into whether and how the bias in the estimates of arm means in adaptive experiments impacts statistical Power and FPR. As shown in Table 4, using the IPW estimator instead of the MLE helps correct the bias of the arm means when data are adaptively collected. In particular, when there is no difference across arms and $n = 785$, we see that the bias in the estimate of arm 1 is only -0.0032 for the IPW estimator, compared to -0.0231 for the MLE estimator. This reduction is also verified when there is a difference between arm means (arm difference 0.1; $p_1 = 0.55$, $p_2 = 0.45$; $n = 785$), where the bias of the IPW estimates for each arm are -0.0005 and

-0.0079 respectively, compared to -0.0031 and -0.1452 respectively for the MLE estimates. Given this reduction in bias, one might expect that we would then see an improvement in the FPR.

**IPW only slightly decreased FPR.** We do see a small reduction in FPR for IPW relative to MLE (11.0% (0.004) vs. 13.4% (0.005)). However, this reduction is not primarily due to the bias reduction per se, as while the bias in arm means is decreased, the bias in the *difference* between the arms means is very similar under both the MLE and IPW estimator: in both cases, the bias is -0.0015. This difference is the numerator of the Wald Z-test, as we focus not on estimating individual arms but on assessing whether the difference between arms is non-zero. Instead, the difference between the tests is in their overall distribution and in the variability of the two test statistics, particularly on the tails. The mass probability we have on the tails of the IPW-adjusted Wald test is lower than the one we have for the standard Wald test (with TS assignment), decreasing FPR, but the mass in the tails is still higher than the 5% probability we would have had with the standard Wald test in a UR experiment, leading to a still elevated FPR.

**IPW greatly reduces Power.** Using IPW slightly reduces FPR (and not fully to the expected 5%) at the cost of decreased Power: relative to the standard Wald test with the MLE, Power decreases from 56% to only 36% for the IPW-adjusted Wald test (see Table 2). The decrease in Power may also be understood based on the distribution of the two test statistics: despite the decreased bias in the estimate of the arm means, there is an increase in the overall variability, which is translated into an increased standard error of the IPW-adjusted Wald test compared to the MLE-based Wald test (standard error of 0.058 vs 0.046, respectively). The median of the distribution of the IPW-adjusted Wald test is 1.27, which is much lower than the median of 2.05 seen for the distribution of the MLE based Wald test (the median is more informative due to the skewness of distributions); a lower median means fewer values exceed the critical values for concluding there is a difference.

**Table 4.** *Estimated arm means, arm differences and absolute arm differences, with their bias and standard errors for the Maximum Likelihood Estimator (MLE) - based both on the Uniform Random (UR - MLE) and Thompson Sampling (TS - MLE) assignment - and the Inverse Probability Weighted (IPW) estimator based on the Thompson Sampling assignment (TS - IPW). We also show the standard error (SE) of the Wald test statistic computed in the three different cases (UR - MLE, TS - MLE; TS - IPW). We report these results for both an arm difference of 0 and of 0.1. Estimates, bias and SE are computed based on 5000 simulated trajectories of size $n = 785$.*

|  | Arm difference 0 $p_1 = p_2 = 0.5$; $n = 785$ | | | Arm difference 0.1 $p_1 = 0.55, p_2 = 0.45$; $n = 785$ | | |
|---|---|---|---|---|---|---|
|  | **UR - MLE** | **TS - MLE** | **TS - IPW** | **UR - MLE** | **TS - MLE** | **TS - IPW** |
| $p_1 - p_2$ estimate | -0.0003 | -0.0015 | -0.0015 | 0.0997 | 0.1421 | 0.1074 |
| $\lvert p_1 - p_2 \rvert$ estimate | 0.0279 | 0.0623 | 0.0015 | 0.0998 | 0.1445 | 0.1074 |
| $p_1$ estimate | 0.5000 | 0.4769 | 0.4967 | 0.5503 | 0.5469 | 0.5494 |
| $p_2$ estimate | 0.4996 | 0.4784 | 0.4982 | 0.4505 | 0.4047 | 0.4420 |
| Bias of $p_1$ estimate | 0.0000 | -0.0231 | -0.0032 | 0.0002 | -0.0031 | -0.0005 |
| Bias of $p_2$ estimate | -0.0003 | -0.0215 | -0.0017 | 0.0005 | -0.1452 | -0.0079 |
| SE of $p_1$ estimate | 0.0003 | 0.0009 | 0.0009 | 0.0003 | 0.0004 | 0.0004 |
| SE of $p_2$ estimate | 0.0003 | 0.0009 | 0.0009 | 0.0003 | 0.0013 | 0.0019 |
| SE of Wald test | 0.0139 | 0.0395 | 0.0428 | 0.0141 | 0.0455 | 0.0583 |
| Median of Distribution of Wald test | 0.0362 | -0.0073 | -0.0028 | 2.8074 | 2.0497 | 1.2727 |

### 5.2 Performance of the TS-induced Wald test

As illustrated at the beginning of this section, the TS-induced Wald test relies on estimating the distribution of the Wald test (under the null) induced by the adaptive nature of the TS algorithm for collecting data. More specifically, assuming there is no difference in arm means ($p_1 = p_2 = 0.5$; n = 785), but data is collected using TS, we simulate what the distribution of the Wald Test would be, and use it to compute the empirical Critical Values for rejecting or not the null-hypothesis (Smith and Villar 2018).

As illustrated in Figure 3, the estimated distribution of the TS-induced Wald test shows a higher variability compared to the standard normal distribution, probably because of the correlation in the reward variable induced by the adaptive algorithm. Thus, the derived Critical Values for getting a FPR control of 5%, are now approximately ±2.6, much higher than the ±1.96 Critical Values of a Wald test distribution in a UR setting. Figure 4 below depicts the logic behind this modification to the statistical test.

**Figure 4:** *This figure shows the principle behind the TS-induced Wald test. The simulated distribution of the Wald Test Statistic (when Arm Difference = 0.0 and assignment is made using Thompson Sampling) can be used to choose the Critical Value in order to get a FPR of 5% (our simulations with p1= 0.5, p2= 0.5; n = 785 lead to Critical Values of ±2.6). However, despite the FPR control, when Arm Difference = 0.1, the Critical Value of ±2.6 produced a Power of 17% only. In contrast, using the "traditional" Critical Values of ±1.96 (chosen for a 5% FPR under UR), led to a 13% FPR and 56% Power. This illustrates clearly how the impact of TS on statistical analysis (and adjustments that impact FPR and Power) cannot merely be handled by changing the test, but suggests additional considerations, e.g., changing the algorithm and data collection strategy.*

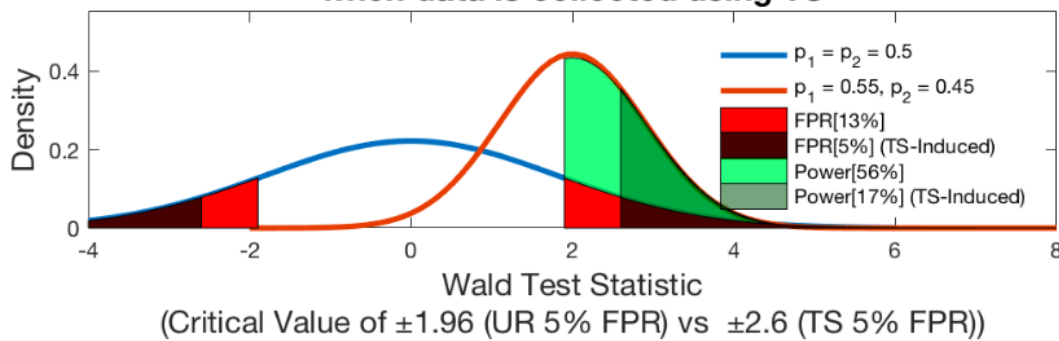

The use of the TS-induced Wald test reduces FPR from 13% to 5%, and crucially, any desired FPR can be achieved, just as with a standard hypothesis test. However, as we can see in row 6 of Table 2, as well as in Figure 4, this comes with a severe cost to Statistical Power: Power is reduced to only 17% for the proposed TS-induced Wald test, as opposed to 56% for standard Wald test with data collected adaptively with TS and 81% for standard Wald test with data collected with UR.

To check the robustness of the proposed strategy, and whether the induced Critical Values would perform well for other instantiations of the null hypotheses, we examined the performance of these new Critical Values when the parameters were actually $p_1 = p_2 = 0.25$ with the same sample size of n = 785. Appendix 3 shows that this only has a minor impact.

### 5.3 Discussion

In summary, both the IPW-adjusted Wald test and the TS-induced Wald Test can reduce the impact of a bandit algorithm on FPR, although both reduce Power. The current work provides empirical illustration that can guide others to take this approach with a range of bandit algorithms and statistical tests. These methods for changing analysis techniques are attractive because they do not require changes to the algorithms' behavior, but simply employ knowledge of how the algorithm behaved. They do not impact participants' experiences, and can be employed anytime after the data are collected. However, the current results also show the statistical analysis problem cannot be easily or readily solved by simply modifying these tests. This suggests the importance of future work that modifies bandit algorithms (or the theoretical frameworks around maximizing reward) to change the data collection strategy, to be more sensitive to statistical considerations like FPR and Power, and the importance of considering whether different strategies are optimal based on the magnitude of the difference in arms.

## 6. Discussion, Limitations & Future Work

Multi-armed bandit algorithms have the potential to be extremely useful for conducting experiments in settings like education and healthcare where there is the potential to assign participants in the experiment to better arms. However, given that the main goal of experiments is to draw conclusions about the arms, such as whether or not they differ from one another, this potential can only be realized if we are confident in our ability to draw correct conclusions about the differences between arms given bandit-collected data. In this paper, we described the use of bandit-driven experiments in three university classes, where instructors appreciated that the experiment had the potential to help their own students but were also concerned about whether the results reflected real differences in the quality of instructional materials. Inspired by this, we conducted an empirical case study demonstrating that Thompson Sampling (TS), a common bandit algorithm, both tends to overestimate differences between arms, with around 70% of simulations showing a greater than 0.7 difference in the posterior mean of the arms when the actual difference in mean was zero, and that it may stop sampling an arm even when there is still significant uncertainty about the arm's mean. These behaviors mean it is difficult to draw conclusions using statistical hypothesis testing: both with a standard Wald test and with Welch's t-test, which corrects for unequal sample sizes, we demonstrated that there were high FPRand lower Power when conducting inference on TS-collected data. This result is not simply due to taking a frequentist approach: analysis of the posteriors using Bayes factors showed similar results. While Inverse Probability Weighting did reduce the bias in the estimated means, this was not enough to correct the increased FPRor decreased Power. Hypothesis testing in which we use knowledge of the TS algorithm to determine the Critical Values for the test statistic does address the heightened FPR, but at a significant cost in Power. As a whole, these results suggest significant barriers to using TS to conduct experiments, despite the fact that it has the potential to benefit participants in the experiment.

While the challenges of statistical hypothesis testing could be construed as only of interest to statisticians or behavioral researchers, they are important to consider from the machine learning perspective both for developing algorithms that can address scientists' real world challenges and for better understanding the behavior of typical bandit algorithms in cases where arms are equivalent. Many of the results we saw in the case study stem from two issues: (a) the tendency of the algorithm to focus on a single arm even when both arms are equivalent, and (b) failing to collect sufficient evidence to rule out the possibility that an arm that appears to be performing badly is in fact reliably worse than another. These tendencies cannot be counteracted solely by changing the way one analyzes the data, since the data look the same in some cases where there truly is a difference between arms and in cases where the arms are equivalent. They also cannot be counteracted solely by adding more participants and thus lengthening the horizon: when the two arms are equivalent, any pattern of sampling is equivalent in terms of regret, and the behavior of focusing on a single arm does not vanish asymptotically. These problems thus suggest that new bandit algorithms are needed if social and behavioral scientists are to take advantage of the potential of experimental designs that use bandits.

The current work is limited in exploring primarily the TS algorithm and considering a limited range of settings, including focusing on binary rewards and two arms. As the setting becomes more complex, we expect these challenges to persist, and further work is needed to explore the tradeoffs between bandits being able to focus on better arms, perhaps permitting experimentation with a larger number of arms, and the inference concerns that we have laid out here. The case studies we developed can provide a foundation for what concerns to consider as well as highlighting the importance of considering cases in which arms are equivalent or nearly equivalent.

In our consideration of hypothesis testing approaches, we focused on illustrating why small changes to the hypothesis testing approach do not address the inference issues satisfactorily. We illustrate how one can integrate machine learning knowledge into how statistical analysis is conducted, showing what works, and what limitations exist. As the more straightforward modifications don't address the FPR and Power issues, this points to the need for further work on developing new statistical tests that consider the assumptions of particular algorithms for data collection (whether bandit algorithms or other methods like best-arm identification). While we believe that changes to algorithms like TS are needed to address these inference issues, there are still promising avenues for collaboration between the machine learning and statistics community to explore how changes in both the test procedures and the data collection might tradeoff between best solutions from a regret standpoint and best solutions from an inference standpoint.

More generally, this paper points to the need for formulating bandit problems that explicitly consider the quality of the evidence collected. Prior work has considered this in some bandit problem formulations, such as best-arm identification (Even-dar et al, 2002; Audibert et al, 2010; Russo 2018), Power-constrained bandit algorithms (Yao et al., 2020), and bandit algorithms that aim to correct for estimation error (Erraqabi et al., 2017). None of these is explicitly concerned, however, with recognizing when arms are equivalent. One potential direction is to consider how to include more uniform randomization when differences between arms are small and thus there is minimal reward to gain. Including more uniform randomization has the potential to not only be beneficial for drawing inferences about whether the arms differ, but also for collecting data that has more value for exploratory analysis.

Real-world experimental settings highlight this need, pointing to a gap in the kinds of problems that are often considered in bandit-related research.

Overall, there remains a great deal for future work to explore that explicitly addresses how to effectively collect data such that participants in experiments are able to benefit from accrued evidence and the collected evidence is such that researchers can draw correct conclusions about the underlying properties of the arms. This work provides both an overview of how and why erroneous conclusions can easily be drawn when data are collected by a bandit algorithm like TS, contrasting the data to that collected via UR sampling or a less effective bandit algorithm (Epsilon-Greedy), and provides a foundation for future work developing bandit algorithms that explicitly consider the data in the objective and that can build on the changes to hypothesis testing techniques that we suggest here.

# Appendix

## Appendix 1. Computation of the Bayes factor in a two-arms binary-reward setting

As discussed in Section 4.4.2, in the context of Bayesian inference, hypothesis testing can be framed as a special case of model comparison where each model refers to a hypothesis. We assume we have two competing hypotheses, each of which corresponds to a separate model: $H_0$ is the model for the null hypothesis, which here is that there is no difference in arm means, and $H_1$ is the model for the alternative hypothesis, which here is that there is some difference in arm means. Bayesian hypothesis testing specifies separate prior probabilities $P(H_0)$ and $P(H_1)$ for each hypothesis. Assuming that we do not have any prior knowledge on which hypothesis may be more plausible, we take $P(H_0) = P(H_1) = 0.5$. Then, based on the observed data $D$ we quantify the evidence in favour of the model $H_0$ and compare it to the evidence in favour of model $H_1$ (or alternatively the evidence against $H_0$). For a given hypothesis $H$, under the Bayesian framework, this evidence is given by the combination of the likelihood function for the observed data, say $P(D|\theta, H)$, which depends on an unknown parameter $\theta$, with each of the prior distributions of the unknown parameter.

We give below details on the likelihood, priors and parameter of interest in our specific two-arms binary setting. For each of the hypothesis-specific models, averaging (i.e., integrating) the likelihood with respect to the prior distribution across the entire parameter space yields the probability of the data under the model and, therefore, the corresponding hypothesis. This quantity is more commonly referred to as the marginal likelihood and represents the average fit of the model to the data. The ratio of the marginal likelihoods for both hypothesis-specific models is known as the Bayes factor, whose general formulation is given by

$$BF_{10} = \frac{P(D|H_1)}{P(D|H_0)} = \frac{\int\limits_{\Theta} P(D|\theta, H_1) P(\theta|H_1) d\theta}{\int\limits_{\Theta} P(D|\theta, H_0) P(\theta|H_0) d\theta} \ ,$$

with $D$ being the observed data, and $\Theta$ the parameter space.

In the context of a two-arm case with binary rewards, the Bayes factor may be computed in closed form by assuming a Binomial model for the reward and a Beta prior for the unknown parameters of the Binomial distribution. More formally, denoting again with $x$ the arm and $r$ the reward, we model the conditional reward for each arm as $r|x = k \sim Binomial\,(1, p_k)$ and assume Beta prior distributions for the unknown parameters $p_k$, with $k = \{1, 2\}$, i.e., $p_k \sim Beta\,(\alpha_k, \beta_k)$. Conjugacy allows us to compute the posterior distribution of the parameter in closed form. After having observed a sample of size $n = n_1 + n_2$, with $n_1$ and $n_2$ the sample size of arm 1 and arm 2, respectively, and $S_1 = \sum_{i=1}^{n_1} r_i$ and $S_2 = \sum_{i=1}^{n_2} r_i$ the number of successes in each group, we have that $p_k | D \sim Beta\,(\alpha_k + S_k, \beta_k + n_k - S_k)$, for $k \in \{1, 2\}$. Here, we consider a non-informative Beta prior distribution for both arms, with $\alpha_1 = \alpha_2 = 1$ and $\beta_1 = \beta_2 = 1$, equivalent to a Uniform distribution in $[0, 1]$.

Comparing the competing hypothesis of no arms difference ($H_0$) and an actual arms difference ($H_1$) requires comparing the marginal likelihood of a model for which the parameters $p_1$ and $p_2$ are the same (thus, the arm means have a common probability $p$, i.e., the success rate distribution does not depend on the arm group) and the marginal likelihood of a model for which the parameters $p_1$ and $p_2$ are different (thus, each arm group has a different success rate). This leads to a pooled prior and then pooled posterior distribution in case of model $H_0$, and to the product of two separate priors and then two separate posterior distribution for model $H_1$. More specifically, we have that $P(D|H_0) = B(\alpha_1 + \alpha_2 + S_1 + S_2, \beta_1 + \beta_2 + n - S_1 - S_2)$ and $P(D|H_1) = B(\alpha_1 + S_1, \beta_1 + n_1 - S_1) \times B(\alpha_1 + S_2, \beta_2 + n_2 - S_2)$, where $B$ denotes the Beta function. Comparing these two quantities will give us the Bayes factor $BF_{10}$, whose value will give us the evidence against the null hypothesis $H_0$, similar to hypothesis testing in the frequentist setting. We choose as "critical" value for "rejecting" the null-hypothesis the cutoff of 3, based on Jeffreys' scales of evidence (Jeffreys, 1961; Kass & Raftery, 1995). If $BF_{10} > 3$, then we chose $H_1$ (and reject $H_0$), otherwise $H_0$.

**Appendix 2. Sensitivity to Priors**

Table 5 presents results comparing Jeffreys' prior (see column Thompson Sampling Jeffreys' Prior) to Beta(1,1) (see column Thompson Sampling Beta(1,1)). Jeffreys' prior is a non-informative prior (Jeffreys 1961) which we use to assess the sensitivity of FPR and Power to the choice of prior. In the Bernoulli reward setting Jeffreys' prior is Beta(½, ½). We see that Jeffreys's prior doesn't influence results substantively: the difference between Beta(1,1) and the Beta(½, ½) Jeffreys' prior with respect to FPR is no greater than 3 % (regardless of hypothesis test). Also, the difference in Power between Beta(1,1) and Jeffreys' prior doesn't differ by more than 5% (regardless of hypothesis test).

      Further prior sensitivity analysis for TS is conducted by Rafferty et al. (Rafferty, Ying, & Williams 2019). In particular, they examine the settings of prior for an arm having a mean which is below, between, and above that of the true arm expected reward in the context of two other statistical test. Priors above encourage more exploration due to being more optimistic, whereas priors below encourage less. The result is that having an optimistic prior will improve Power and reduce FPR somewhat.

**Table 5**: *Comparing False Positive Rate and Power for Beta(1,1) and Beta(½, ½) priors for Thompson Sampling, with various hypothesis tests. FPR and Power for analysis of data collected using TS with Beta(1,1) prior and Thompson Sampling with Beta(½ ,½) Jeffreys' prior. Each row shows FPR and Power for a different Statistical Analysis, including: (1) Wald test, (2) Welch's t-test used for unequal sample sizes or variances, (3) Bayes factor (with cutoff 3.0). (4) IPW-adjusted Wald test: Wald test where the sample means are adjusted using the TS Assignment Probabilities to do Inverse Probability Weighting (IPW). (5) TS-induced Wald test: Wald test where a non-parametric distribution is used that is generated by simulations of running TS under the null hypothesis that arm means are equal. The Standard Errors (SE) are written in parenthesis.*

| | Thompson Sampling Beta(1, 1) | | Thompson Sampling Jeffreys' Prior Beta(½, ½) | |
|---|---|---|---|---|
| **Statistical Test** | FPR (SE) [ $n = 785$ ; | Power (SE) [ $n = 785$ ; | FPR (SE) [ $n = 785$ ; | Power (SE) [ $n = 785$ ; |

| | $p_1 = p_2 = 0.5$ ] | $p_1 = 0.55$ $p_2 = 0.45$ ] | $p_1 = p_2 = 0.5$ ] | $p_1 = 0.55$ $p_2 = 0.45$ ] |
|---|---|---|---|---|
| 1. Wald test | 13.3 % (0.5) | 56.2 % (0.7) | 14.4 % (0.5) | 56.1 % (0.7) |
| 2. Welch's t-test | 11.9 % (0.5) | 52. 4% (0.7) | 12.3 % (0.5) | 52.2 % (0.7) |
| 3. Bayes factor (Cutoff 3.0 ) | 4.2 % (0.3) | 19.3 % (0.6) | 4.9 % (0.3) | 21.4 % (0.6) |
| 3. Bayes factor (Cutoff 1.0) | 9.3 % (0.4) | 43.5 % (0.7) | 10.5 % (0.4) | 44.1 % (0.7) |
| 4. Bayes factor (Cutoff 0.4) | 18.6 % (0.6) | 69.5 % (0.7) | 19.2 % (0.6) | 69.1 % (0.7) |
| 5. IPW-adjusted Wald test | 11.0 % (0.4) | 35.9 % (0.7) | 12.6 % (0.5) | 37.8 % (0.7) |
| 6. TS-induced Wald Test | 4.5 % (0.3) | 17.4 % (0.5) | 4.2 % (0.3) | 14.4 % (0.5) |

### Appendix 3. Comparing other Null Hypothesis settings

When arm means are equal, the main text focuses on a case where $p_1 = p_2 = 0.50$. Here, we examine the consequences of alternative probabilities, focusing on $p_1 = p_2 = 0.25$, in order to assess whether the impact on FPR is relatively constant across varying instantiations of the null hypothesis. Comparing hypothesis testing with data from $p_1 = p_2 = 0.5$ (Table 2) versus hypothesis testing with data from p1=p2=0.25 shown in Table 6 demonstrates that results are very similar: for Wald's test, Welch's test, the Bayes' Factor test, and the IPW-adjusted Wald test no False Positive Rates differ by more than 1%.

For the TS-induced Wald test, we make two comparisons to assess the impact of using $p_1 = p_2 = 0.25$ rather than $p_1 = p_2 = 0.50$. First, in row 5 of Table 6, we perform our hypothesis test with cutoffs derived from simulating with $p_1 = p_2 = 0.25$. We again see relatively similar performance to the equivalent version in the main text: a FPR of 5.2% for $p_1 = p_2 = 0.25$ compared to 4.5% for $p_1 = p_2 = 0.50$.

In both of these cases, we have assumed that the values for the arm mean that is assumed by the algorithm-induced cutoff matches the true values of the arm mean. However, this information is not known to the experimenter ahead of time, and thus the TS-induced cutoffs may be set using a value that doesn't match the true rewards, even if some attempts are made to estimate the true values such as by taking the overall average reward across all samples. Thus, we show the results of Wald's test with TS-induced cutoffs when the cutoff simulations are based an incorrect arm mean; in particular, we chose the cutoffs based on an assumption that $p_1 = p_2 = 0.50$, but in actuality, $p_1 = p_2 = 0.25$. We find that FPR is still well-controlled, with a value of 5.0% (row 6 of Table 6).

**Table 6**: *Examining False Positive Rate with $p_1 = p_2 = 0.25$, with various hypothesis tests. FPR using TS, Uniform Random, Epsilon Greedy 0.1. Each row shows FPR for a different Statistical Analysis. Rows 5 and 6 vary the assumption of arm values: row 5 matches the true values of the arms, while row 6 has cutoffs chosen by assuming the true reward probabilities are $p_1 = p_2 = 0.50$. The Standard Errors (SE) are in parenthesis.*

| | FPR (SE) $[\, n = 785 \,; p_1 = p_2 = 0.25 \,]$ | | |
|---|---|---|---|
| | **Thompson Sampling** | **Uniform Random** | **Epsilon Greedy 0.1** |
| 1. Wald test | 13.7 (0.5) % | 5.6 (0.3) % | 6.0 (0.3) % |
| 2. Welch's t-test | 12.6 (0.5) % | 5.5 (0.3) % | 5.4 (0.3) % |
| 3. Bayes Factor (Cutoff 3.0 ) | 4.3 % (0.3) | 0.5 % (0.1) | 1.3 % (0.2) |
| 3. Bayes factor (Cutoff 1.0) | 10.0 % (0.4) | 1.9 % (0.2) | 3.7 % (0.3) |
| 4. Bayes factor (Cutoff 0.4) | 19.1 % (0.6) | 5.7 % (0.3) | 12.2 % (0.5) |
| 5. IPW-adjusted Wald test | 10.8 (0.4) % | | |
| 6. TS-induced Wald test 0.25 | 5.2 (0.3) % | | |
| 7. TS-induced Wald test 0.5 | 5.0 (0.3) % | | |

## Appendix 4. Welch's t-test or Unequal Variances t-test

We compare our results when using the Welch's t-test (Welch, 1947). Welch's t-test consist of hypothesis testing based on the following test statistic

$$\frac{\bar{Y}_2 - \bar{Y}_1}{\sqrt{\frac{s_1^2}{n_1} + \frac{s_2^2}{n_2}}},$$

where $s_k$ is the sample standard deviation, $Y_k$ is the sample mean, and $n_k$ is the sample size for arm k. The Welch's t-test accommodates unequal sample distribution variance. The distribution of the test-statistic is a t-distribution with $\nu$ degrees of freedom, where $\nu$ is given as

$$\nu = \frac{(s_1{}^2 + s_2{}^2)^2}{\frac{s_1{}^4}{N_1{}^2 v_1} + \frac{s_2{}^4}{N_2{}^2 v_2}}$$

With $\nu_{1 = N_1 - 1}$, $\nu_2 = N_2 - 1$. With larger sample sizes, approaching n of 100 or more, the tests become equivalent – and are equivalent asymptotically.

## Appendix 5. Mean Reward

In Table 7 we compare mean reward for TS, Thompson Sampling Jeffreys' Prior, Epsilon-Greedy (epsilon = 0.1), and UR policies for sample size $n = 785$, and with an arm difference of 0.1 ($p_1 = 0.45$, $p_2 = 0.55$ ). Mean reward is computed as the mean reward for a given sample size, averaged over 5000 simulations. We see that UR achieves the lowest mean reward of 0.5, Epsilon-Greedy (epsilon = 0.1) achieves greater mean reward than UR, but lower than TS with or without Jeffreys' prior (0.532 vs. 0.536).

**Table 7**: *Mean Reward. Mean reward averaged over 5000 simulations is shown for Thompson Sampling, Thompson Sampling Jeffreys' Prior, Epsilon-Greedy (epsilon = 0.1), and Uniform Random policies. Standard Errors (SE) are shown in brackets. Results are shown for sample size $n = 785$, and with an arm difference of 0.1 ($p_1 = 0.45$, $p_2 = 0.55$ ).*

| Mean Reward (SE) [ $n = 785$ ; $p_1 = 0.45$, $p_2 = 0.55$ ] | | | |
|---|---|---|---|
| **Thompson Sampling** | **Thompson Sampling Jeffreys' Prior** | **Epsilon-Greedy (epsilon = 0.1)** | **Uniform Random** |
| 0.536 (0.0) | 0.536 (0.0) | 0.532 (0.0) | 0.5 (0.0) |